# Distributed Learning based on 1-Bit Gradient Coding in the Presence of Stragglers

Chengxi Li, Mikael Skoglund, *Fellow, IEEE*

*Abstract*—This paper considers the problem of distributed learning (DL) in the presence of stragglers. For this problem, DL methods based on gradient coding have been widely investigated, which redundantly distribute the training data to the workers to guarantee convergence when some workers are stragglers. However, these methods require the workers to transmit real-valued vectors during the process of learning, which induces very high communication burden. To overcome this drawback, we propose a novel DL method based on 1-bit gradient coding (1-bit GC-DL), where 1-bit data encoded from the locally computed gradients are transmitted by the workers to reduce the communication overhead. We theoretically provide the convergence guarantees of the proposed method for both the convex loss functions and non-convex loss functions. It is shown empirically that 1-bit GC-DL outperforms the baseline methods, which attains better learning performance under the same communication overhead.

## I. INTRODUCTION

IN ORDER to take full advantages of edge computational resources, distributed learning (DL) has attracted much attention [1-8]. In a typical setting of DL, a large training dataset is partitioned and distributed among a number of workers, and the aim is to minimize a loss function with the training dataset by optimizing the parameter vector [9-11]. In each iteration of the learning process, each worker obtains a new update vector of the parameter vector based on its local training data and then broadcasts this update vector to other workers. After that, each worker aggregates the local update vectors received from other workers to obtain the global update vector and updates the parameter vector [12].

In practice, some workers may be unresponsive occasionally caused by unexpected incidents such as

C. Li and M. Skoglund are with the Division of Information Science and Engineering, School of Electrical Engineering and Computer Science, KTH Royal Institute of Technology, 10044 Stockholm, Sweden (e-mail: chengxli@kth.se; skoglund@kth.se). Corresponding author: Chengxi Li.

hardware failures and network congestion, which are known as stragglers [13-15]. Under the framework of DL, stragglers may significantly degrade the convergence performance. To mitigate the negative impact of stragglers, DL methods based on gradient coding (GC) have been proposed, which replicate each training data sample, send it to multiple workers, and let the workers transmit vectors encoded from the locally computed gradients [16-22]. While GC implies increasing computation load and storage at the workers, it can speed up convergence by making it possible to compute an exact or approximate version of the global update vector to update the parameter vector, even when the stragglers are present.

The DL methods based on GC can be roughly categorized into two classes, i.e., DL methods based on exact GC and DL methods based on approximate GC, based on whether they compute an exact or approximate version of the global update vector. In DL methods based on exact GC [16-18], even with the presence of some stragglers, the same global update vector can be obtained and utilized as if there were no stragglers. The original work of exact GC is conducted in [16], where the training data are assigned to the workers based on fractional repetition codes (FRC) and the exact global update vector can be recovered with worst-case stragglers. In [16], each worker transmits a linear combination of gradients without any knowledge about who the stragglers will be, and the linear combinations are designed so that the span of the vectors sent by any fixed number of non-straggler workers contains the exact global gradient. Later, to improve the performance of the method in [16], GC with static clustering is proposed in [17], where the workers are divided into clusters and GC is applied at the cluster level. Different from [17], in [18] GC with dynamic clustering is proposed, where the clusters are dynamically formed among the iterations. Motivated by the observation that DL is robust to noise [23], in many practical settings, it is sufficient to recover the global update vector approximately to update the parameter vector and obtaining an exact version of the global update vector is not necessary. Based on this, DL methods based on approximate GC have been investigated [20-22]. In [20], an approximate GC technique is designed based on expander graphs and the decoding error is analyzed when using the optimal decoding coefficients. In [21], an inexact gradient is recovered in each iteration based on FRC and a rigorous analysis in terms of the convergence properties is provided. In [22], a DL method based on stochastic gradient coding (SGC-DL) is proposed, which works with random stragglers and distributes the training dataset to workers in a pair-wise balanced manner. Among these DL methods based on approximate GC, the merits of SGC-DL lie in its simplicity of decoding and requiring only a small amount of redundancy of data distribution to mitigate the impact of a huge number of stragglers. However, SGC-DL has obvious shortcoming. During each iteration, the workers send real-valued vectors, which induces very high communication overhead.

DL requires frequent communication among the workers for exchanging information to update the parameter vector and complete the learning process. Under this framework, the communication overhead is a key factor that the system designer must take into account, especially in wireless networks with very limited bandwidth [9, 24-28, 41-46]. Various schemes of communication compression have been proposed to decrease the communication cost of DL, which reduce the data transmitted by the workers by means of gradient quantization or gradient sparsification. In gradient sparsification, elements in the gradient vectors with significant magnitudes are selectively transmitted based on a given threshold. For example, in [29], a general gradient sparsification (GGS) method is proposed. This method adopts gradient correction to enhance convergence performance by addressing the accumulated insignificant gradients in a more appropriate manner. In [30], a sparse binary compression (SBC) method is introduced, which merges techniques from communication delay and gradient sparsification to achieve a better trade-off between gradient sparsity and temporal sparsity. High compression rate of communication can be achieved by gradient sparsification techniques. However, it is at the cost of severely degraded convergence performance. As an alternative, in gradient quantization, elements in the gradient vectors are quantized to fewer bits and the workers only transmit quantized vectors instead of real-valued version of them. For instance, in [31], the authors propose a federated trained ternary quantization (FTTQ) method, where the quantized networks are optimized at the workers with a self-learning quantization factor. In [32], a framework of hierarchical gradient quantization is studied, whose convergence rate is analyzed as a function of quantization bits. Among existing gradient quantization schemes, 1-bit DL methods are becoming increasingly popular and appealing due to their simple compression rule and extremely low communication cost per iteration, which allow workers to transmit only 1-bit data in the DL system [33, 34, 37]. However, the existing 1-bit DL methods do not take stragglers into consideration, leading to a significant performance drop when stragglers are present. To the best of our knowledge, 1-bit quantization techniques have not been thoroughly employed to tackle DL problems in scenarios with stragglers in order to reduce the communication overhead.

In this paper, for the problem of DL with stragglers, to reduce the communication overhead of SGC-DL, we propose a new DL method based on 1-bit gradient coding (1-bit GC-DL). The proposed method consists of two stages. In the **data distribution stage**, the training data are distributed redundantly to the workers in a pair-wise balanced manner, as done in [22]. In the **learning stage**, in each iteration, each non-straggler worker computes the gradients of all the training data samples allocated to it and obtains a weighted sum of the gradients. After that, each non-straggler worker performs random 1-bit quantization of the weighted sum and broadcasts the 1-bit vectors to its peers. The workers receive the 1-bit vectors and update the parameter vector based on them. For 1-bit GC-DL, we conduct convergence analysis when the loss functions are convex or non-convex. Using numerical results, we show that 1-bit GC-DL outperforms the baseline methods and attains better learning performance under the same communication overhead. The main contributions of this paper are given as follows:

1) We propose a new method, i.e., 1-bit GC-DL, to reduce the communication overhead of DL in the presence of stragglers by combining the advantages of GC and 1-bit quantization;

2) We conduct rigorous convergence analysis of 1-bit GC-DL for different types of loss functions;

3) We provide empirical evidence to verify that 1-bit GC-DL outperforms the existing methods in terms of the learning performance under the same communication overhead.

It is worth noting that although 1-bit quantization techniques have been widely used in the existing DL methods [33, 34, 37], there are significant differences between our work and these existing methods, which are listed as follows:

1) Implementation Differences: The existing 1-bit DL methods conduct quantization based on individual gradient vectors at each worker. In contrast, our proposed method employs 1-bit quantization on the weighted sum of multiple local gradients, where each local gradient may be redundantly used across several workers based on the GC strategy. Accordingly, the aggregation rules for 1-bit vectors from different workers are distinct between our method and existing ones, since 1-bit vectors are generated in totally different ways.

2) Theoretical Analysis: Owing to the differences in implementation, our theoretical analysis and convergence performance significantly differ from those in existing works.

3) Performance: Our method effectively addresses the issue of stragglers, unlike existing 1-bit DL methods. When stragglers are present, the performance of existing methods degrades significantly. In contrast, our proposed method mitigates the negative impact of stragglers by combining the advantages of GC and 1-bit quantization.

The rest of this paper is organized as follows. In Section II, the problem model is formulated. In Section III, the proposed method is introduced. In Section IV, we provide theoretical analysis of the proposed method. In Section V, numerical results are shown to demonstrate the superiority of the proposed method. In Section VI, we draw the conclusion.

## II. PROBLEM MODEL

There is a training dataset composed of $m$ training data samples, denoted by $\mathbf{A} \in \mathbb{R}^{m \times l}$, where $\mathbf{a}_i \in \mathbb{R}^l$ is the $i-$th row of $\mathbf{A}$ and represents the $i-$th data sample of length $l$. The learning problem can be formulated as an optimization problem [22]:

$$\boldsymbol{\beta}^* = \arg\min_{\boldsymbol{\beta}} \hat{L}(\mathbf{A}, \boldsymbol{\beta}), \qquad (1)$$

where $\boldsymbol{\beta} \in \mathbb{R}^w$ is the parameter vector, and $\hat{L}(\mathbf{A}, \boldsymbol{\beta})$ is the loss function defined as

$$\hat{L}(\mathbf{A}, \boldsymbol{\beta}) = \sum_{i=1}^m L(\mathbf{a}_i, \boldsymbol{\beta}). \qquad (2)$$

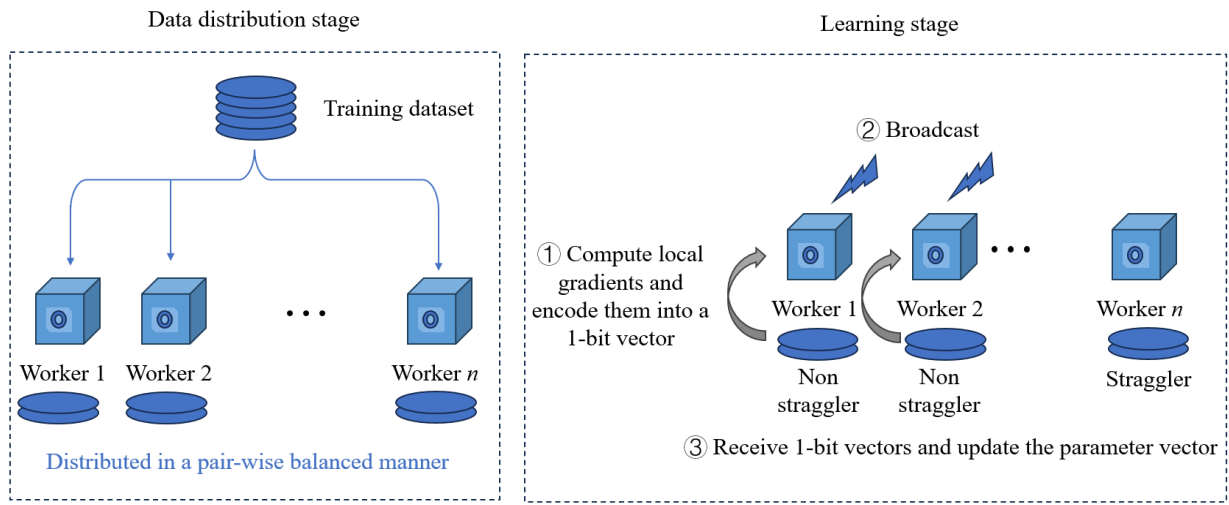

Fig. 1. The flowchart of 1-bit GC-DL.

In (2), $L(\mathbf{a}_i, \boldsymbol{\beta})$ is the training loss associated with data sample $\mathbf{a}_i$, whose form can vary depending on the specific problem at hand, such as regression problems.

Our aim is to deal with the problem in (1) in a DL system consisting of $n$ workers, where solving the problem involves two stages, i.e., the data distribution stage and the learning stage [22]:

1) In the **data distribution stage**, data sample $\mathbf{a}_i$ is distributed to $d_i$ workers, $i = 1, ..., m$. For the $j-$th worker, set $\mathcal{S}_j$ contains all the indices of the data samples that are assigned to it, $j = 1, ..., n$.

2) The **learning stage** is made up of iterations. It is assumed that, in each iteration, the probability of each worker to be a straggler is $p$, and the straggler behavior is independent among workers and iterations. Similar assumptions have also been made in [20, 22]. In each iteration, if the $j-$th worker is not a straggler, it computes the gradients corresponding to the data samples in $\mathcal{S}_j$, encodes the gradients into an update vector and broadcasts information to its peers [12]. After that, each worker receives information from other non-straggler workers, aggregates the information and updates the parameter vector accordingly.

For this problem, SGC-DL proposed in [22] allocates the training data samples to the workers in a pair-wise balanced manner in the data distribution stage. After that, in the learning stage, in each iteration, each non-straggler worker encodes the locally computed gradients based on the local training data samples into a real-valued update vector and broadcasts it to its peers. Meanwhile, each worker receives the real-valued update vectors from others and utilizes it to update the parameter vector. Although SGC-DL provides convergence guarantees, it induces very high communication burden when transmitting real-valued vectors in the system. To reduce the communication overhead of SGC-DL, we will propose a new method in Section III. It is worth noting that, our primary focus lies in the communication cost among workers in the learning stage instead of the communication cost induced by distributing training data samples in the data distribution stage. This emphasis is based on the observation that inter-worker communication often presents more significant bottlenecks in practical applications. This fact is also suggested by [16, 19] where the communication cost associated with distributing training samples prior to training is not considered.

## III. THE PROPOSED METHOD

In this section, we introduce the proposed method to address the problem in Section II and to overcome the drawback of SGC-DL. The flowchart of the proposed method is shown in Fig. 1.

In the **data distribution stage**, the training data samples are distributed to the workers in a pair-wise balanced manner, as done in [22]. In this scheme, data sample $\mathbf{a}_i$ is distributed to $d_i$ workers and the number of workers that hold both $\mathbf{a}_i$ and $\mathbf{a}_{i'}$ is $d_i d_{i'}/n$ for $i \neq i'$. This approach is favored for its tractability in theoretical analysis and its ability to closely approximate a completely random distribution of the training data samples, where $d_i$ workers are selected randomly and independently for each training sample.

In the **learning stage**, in the $t-$th iteration, we use an indicator function as follows to show whether the $j-$th worker is a straggler:

$$I_j^t = \begin{cases} 1, & \text{if the } j\text{-th worker is not a straggler,} \\ 0, & \text{if the } j\text{-th worker is a straggler.} \end{cases} \quad (3)$$

In the $t-$th iteration, if $I_j^t = 1$, the $j-$th worker computes the gradients corresponding to the data samples it holds based on the current parameter vector $\boldsymbol{\beta}_t$, which yields $\left\{ \nabla L(\mathbf{a}_i, \boldsymbol{\beta}_t), i \in \mathcal{S}_j \right\}$. After that, the local gradients are encoded into a 1-bit vector $\mathbf{h}_j^t$ of length $w$, and its $k-$th element can be expressed as

$$\left( \mathbf{h}_j^t \right)_k = Q\left[ \left( \mathbf{f}_j^t \right)_k \right], j \in \left\{ j \mid I_j^t = 1 \right\}, \quad (4)$$

where

$$\mathbf{f}_j^t = \sum_{i \in \mathcal{S}_j} \frac{1}{d_i(1-p)} \nabla L(\mathbf{a}_i, \boldsymbol{\beta}_t), \quad (5)$$

set $\mathcal{S}_j$ contains all the indices of the data samples that are assigned to the $j-$th worker, $j = 1, ..., n$, and $Q(\cdot)$ is the random 1-bit quantization function. In (4), the operation of $Q(\cdot)$ on the $k-$th element of $\mathbf{f}_j^t$ can be expressed as

$$Q\left[ \left( \mathbf{f}_j^t \right)_k \right] \triangleq \begin{cases} +1, & \text{with probability } \dfrac{1}{2} + \dfrac{1}{2} \dfrac{\left( \mathbf{f}_j^t \right)_k}{\left\| \mathbf{f}_j^t \right\|}, \\ -1, & \text{with probability } \dfrac{1}{2} - \dfrac{1}{2} \dfrac{\left( \mathbf{f}_j^t \right)_k}{\left\| \mathbf{f}_j^t \right\|}, \end{cases} \quad (6)$$

where $\left\| \mathbf{f}_j^t \right\|$ is the Euclidean norm of $\mathbf{f}_j^t$. Technical importance of $Q(\cdot)$ in (6) lies in the fact that, different from the deterministic 1-bit quantization operation, $Q(\cdot)$ is unbiased with a scaling factor $\left\| \mathbf{f}_j^t \right\|$ [34]. Upon doing this, the $j-$th non-straggler worker broadcasts $\mathbf{h}_j^t$ and $\left\| \mathbf{f}_j^t \right\|$ to its peers. After receiving the information from others, each worker updates the parameter vector in the following way:

$$\boldsymbol{\beta}_{t+1} = \boldsymbol{\beta}_t - \gamma_t \hat{\mathbf{g}}_t, \quad (7)$$

where $\gamma_t$ is the learning rate and $\hat{\mathbf{g}}_t$ is the global update vector defined as

$$\hat{\mathbf{g}}_t \triangleq \sum_{j=1}^n I_j^t \mathbf{h}_j^t \left\| \mathbf{f}_j^t \right\|. \quad (8)$$

The key novelty of the proposed method is that the local gradients are encoded into 1-bit vectors based on random 1-bit quantization and then transmitted by the non-straggler workers. In contrast, SGC-DL encodes the local gradients into real-valued vectors through weighted sum. Let us denote $\rho$ as the number of bits transmitted by each non-straggler worker in each iteration. It is obvious that a larger value of $\rho$ indicates higher communication burden per iteration. For the proposed method, we have $\rho_{\text{1-bit GC-DL}} = w + \zeta$, where $\zeta$ is the number of bits needed to transmit real-valued $\left\| \mathbf{f}_j^t \right\|$. For SGC-DL, we have

$\rho_{\text{SGC-DL}} = w\zeta$. Considering that the parameter vector is always high-dimensional, i.e., $w \gg \zeta$, we can derive that $\rho_{\text{SGC-DL}} \gg \rho_{\text{1-bit GC-DL}}$. Based on that, the communication overhead per iteration can be significantly reduced by the use of 1-bit GC-DL. It is also worth noting that the differing encoding techniques between SGC-DL and 1-bit GC-DL necessitate distinct convergence analyses, which we will detail in Section IV.

## IV. Theoretical Analysis

In this section, we conduct theoretical analysis for 1-bit GC-DL, which provides rigorous convergence guarantees under different settings. To be more specific, we consider two types of loss functions, i.e., strongly convex loss functions and non-convex loss functions.

### A. $\lambda$-strongly convex loss functions

We first analyze the convergence performance of the proposed method under the case where the loss function is $\lambda$-strongly convex. The definition of $\lambda$-strongly convex is given as follows.

*Definition 1 (Strongly Convex):* The loss function $\hat{L}$ is $\lambda$-strongly convex if

$$\hat{L}(\mathbf{A}, \boldsymbol{\beta}) \geq \hat{L}(\mathbf{A}, \boldsymbol{\beta}') + \left\langle \nabla \hat{L}(\mathbf{A}, \boldsymbol{\beta}'), \boldsymbol{\beta} - \boldsymbol{\beta}' \right\rangle + \frac{\lambda}{2} \left\| \boldsymbol{\beta} - \boldsymbol{\beta}' \right\|^2, \forall \boldsymbol{\beta}, \boldsymbol{\beta}', \quad (9)$$

where $\langle \cdot, \cdot \rangle$ denotes the inner product of two vectors.

Before providing a theorem which describes the convergence performance of the proposed method, we first provide three lemmas to aid the derivation of the theorem.

*Lemma 1:* The global update vector $\hat{\mathbf{g}}_t$ defined in (8) is an unbiased estimator of $\nabla \hat{L}(\mathbf{A}, \boldsymbol{\beta}_t)$.

*Proof:* Let us denote $\mathbb{E}\left( \cdot \mid \mathcal{F}_{0:t} \right)$ as the expectation taken over the random behavior of the stragglers in the $t-$th iteration conditioned on the previous iterations. Then, based on (8), we have

$$\mathbb{E}\left( \hat{\mathbf{g}}_t \mid \mathcal{F}_{0:t} \right) = \mathbb{E}\left( \sum_{j=1}^n I_j^t \mathbf{h}_j^t \left\| \mathbf{f}_j^t \right\| \mid \mathcal{F}_{0:t} \right) = \sum_{j=1}^n \mathbb{E}\left( I_j^t \mathbf{h}_j^t \left\| \mathbf{f}_j^t \right\| \mid \mathcal{F}_{0:t} \right)$$
$$\overset{\langle 1 \rangle}{=} \sum_{j=1}^n \mathbb{E}\left( I_j^t \mid \mathcal{F}_{0:t} \right) \mathbb{E}\left( \mathbf{h}_j^t \left\| \mathbf{f}_j^t \right\| \mid \mathcal{F}_{0:t} \right) = \sum_{j=1}^n (1-p) \mathbf{f}_j^t, \quad (10)$$

where $\langle 1 \rangle$ is derived from the conditional independence between the straggler behavior of the workers and the random 1-bit quantization. Next, substituting (5) into (10) yields

$$\mathbb{E}\left(\hat{\mathbf{g}}_t | \mathcal{F}_{0:t}\right) = \sum_{j=1}^{n}(1-p)\sum_{i\in\mathcal{S}_j}\frac{1}{d_i(1-p)}\nabla L(\mathbf{a}_i,\boldsymbol{\beta}_t)$$

$$= \sum_{j=1}^{n}\sum_{i=1}^{m}\frac{I\left(i\in\mathcal{S}_j\right)}{d_i}\nabla L(\mathbf{a}_i,\boldsymbol{\beta}_t)$$

$$= \sum_{i=1}^{m}\frac{\nabla L(\mathbf{a}_i,\boldsymbol{\beta}_t)}{d_i}\sum_{j=1}^{n}I\left(i\in\mathcal{S}_j\right) \qquad (11)$$

$$= \sum_{i=1}^{m}\nabla L(\mathbf{a}_i,\boldsymbol{\beta}_t) = \nabla\hat{L}(\mathbf{A},\boldsymbol{\beta}_t),$$

where $I\left(i\in\mathcal{S}_j\right)=1$ if $i\in\mathcal{S}_j$ and $I\left(i\in\mathcal{S}_j\right)=0$ otherwise. ∎

*Lemma 2:* Suppose that the gradients are all bounded by a constant[1]:

$$\left\|\nabla L(\mathbf{a}_i,\boldsymbol{\beta}_t)\right\|^2 \le C, \forall i, \forall \boldsymbol{\beta}_t, \qquad (12)$$

then $\mathbb{E}\left(\left\|\hat{\mathbf{g}}_t\right\|^2\right)$ is upper bounded by

$$\mathbb{E}\left(\left\|\hat{\mathbf{g}}_t\right\|^2\right) \le$$

$$Cm^2 + \left[w-(1-p)\right]\frac{m^2-m}{n(1-p)}C + \frac{\left[w-(1-p)\right]}{(1-p)}C\sum_{i=1}^{m}\frac{1}{d_i}. \qquad (13)$$

*Proof:* Please see Appendix A. ∎

*Lemma 3 (Lemma 1 in [35]):* Suppose $\hat{L}$ is a $\lambda$-strongly convex function and that $\hat{\mathbf{g}}_t$ is an unbiased estimator of the gradient of $\hat{L}$ at $\boldsymbol{\beta}_t$ with $\mathbb{E}\left(\left\|\hat{\mathbf{g}}_t\right\|^2\right)\le G$. If the learning rate is set as $\gamma_t = 1/(\lambda t)$ and the parameter vector is updated iteratively as (7), for any $T$ it holds that

$$\mathbb{E}\left(\left\|\boldsymbol{\beta}_T - \boldsymbol{\beta}^*\right\|^2\right) \le \frac{4G}{\lambda^2 T}. \qquad (14)$$

*Theorem 1 (Convergence Performance of 1-Bit GC-DL for $\lambda$-Strongly Convex Loss Functions):* Suppose that the loss function $\hat{L}$ is $\lambda$-strongly convex and $\left\|\nabla L(\mathbf{a}_i,\boldsymbol{\beta}_t)\right\|^2$ is upper bounded as (12), then by setting $\gamma_t = 1/(\lambda t)$, it holds for the proposed method that

$$\mathbb{E}\left(\left\|\boldsymbol{\beta}_T - \boldsymbol{\beta}^*\right\|^2\right) \le$$

$$\frac{4\left\{Cm^2 + \left[w-(1-p)\right]\frac{m^2-m}{n(1-p)}C + \frac{\left[w-(1-p)\right]}{(1-p)}C\sum_{i=1}^{m}\frac{1}{d_i}\right\}}{\lambda^2 T}. \qquad (15)$$

*Proof:* Combining Lemma 1, Lemma 2 and Lemma 3 yields Theorem 1 straightforwardly. ∎

From Theorem 1, it can be seen that the convergence of 1-bit GC-DL can be accelerated by increasing the values of $d_i$, which is at the cost of increasing computation load and storage at the workers[2]. This is consistent with our intuition. Let us define the average redundancy of data distribution as [22]

$$\bar{d} = \frac{1}{m}\sum_{i=1}^{m}d_i, \qquad (16)$$

which indicates how redundant the training data are distributed to the workers in an average sense. When $\bar{d}$ is fixed, we are interested in investigating how to assign values to $\{d_i, \forall i\}$ in order to minimize the upper bound in (15) and to attain optimal convergence performance. This can be equivalently expressed as the following optimization problem:

$$\min_{\{d_i,\forall i\}} \frac{4\left\{Cm^2 + \left[w-(1-p)\right]\frac{m^2-m}{n(1-p)}C + \frac{\left[w-(1-p)\right]}{(1-p)}C\sum_{i=1}^{m}\frac{1}{d_i}\right\}}{\lambda^2 T},$$

$$\text{s.t. } \bar{d} = D, \qquad (17)$$

which is equivalent to

$$\min_{\{d_i,\forall i\}} \sum_{i=1}^{m}\frac{1}{d_i}, \text{ s.t. } \bar{d} = D. \qquad (18)$$

In (17) and (18), $D$ is a constant, which indicates the average redundancy of data distribution is fixed. The solution to (18) is very straightforward and can be given as

$$d_i = D, \forall i. \qquad (19)$$

The solution in (19) indicates that the optimal convergence performance can be attained when each training data sample is distributed to the same number of workers. Hereinafter, we will refer to this scheme as the **homogeneous data distribution scheme**.

### B. Non-convex loss functions

Next, we analyze the convergence performance of the proposed method under the case where the loss function is non-convex and smooth. The definition of smooth function is given as follows.

*Definition 2 (Smooth):* The loss function $\hat{L}$ is smooth with a constant $S \ge 0$ if

$$\hat{L}(\mathbf{A},\boldsymbol{\beta}) \le \hat{L}(\mathbf{A},\boldsymbol{\beta}') + \left\langle\nabla\hat{L}(\mathbf{A},\boldsymbol{\beta}'),\boldsymbol{\beta}-\boldsymbol{\beta}'\right\rangle + S\left\|\boldsymbol{\beta}-\boldsymbol{\beta}'\right\|^2, \forall\boldsymbol{\beta},\boldsymbol{\beta}'. \qquad (20)$$

In the following analysis, the homogeneous data distribution scheme in (19) is adopted due to its optimality demonstrated in Section IV.A for strongly convex loss functions. We state the convergence performance of 1-bit GC-DL when using constant learning rates and decaying learning rates in Theorem 2 and Theorem 3, respectively.

*Theorem 2 (Convergence Performance of 1-Bit GC-DL for Non-Convex Loss Functions with Constant Learning*

---

[1]The assumption of bounded gradients is a common practice in the field of DL, as evidenced by its application in several foundational studies including [22, 47].

[2]In this paper, we consider scenarios where the straggler behavior of the workers is caused by unexpected incidents, such as hardware failures and network congestion. These incidents are independent of the computation load at each worker. In other words, the probability of a worker becoming a straggler in each iteration is not influenced by its computation load. Analysis under the cases where the probability of a worker becoming a straggler varies with the computation load across different workers are left for future work.

*Rates):* Suppose that the gradients are all bounded by (12), 1-bit GC-DL with constant learning rates

$$\gamma_t = \gamma = \frac{1 - \sqrt{1 - 4S / (T+1)^{3/4}}}{2S}, \quad (21)$$

converges as follows:

$$\frac{1}{T+1} \sum_{t=0}^{T} \mathbb{E}\left( \left\| \nabla \hat{L}(\mathbf{A}, \boldsymbol{\beta}_t) \right\|^2 \right)$$

$$\leq \frac{L^0 - L^*}{(T+1)^{1/4}}$$

$$+ (T+1)^{3/4} \left( \frac{1 - \sqrt{1 - 4\dfrac{S}{(T+1)^{3/4}}}}{2S} \right)^2 \frac{w - (1-p)}{1-p} \left( \frac{m-1}{n} + \frac{1}{D} \right) CmS,$$

for $T > (4S)^{4/3} - 1$,

$$\quad (22)$$

where $L^0 = \hat{L}(\mathbf{A}, \boldsymbol{\beta}_0)$ and $L^* = \min_{\boldsymbol{\beta}} \hat{L}(\mathbf{A}, \boldsymbol{\beta})$ . Based on (22), we have

$$\lim_{T \to +\infty} \frac{1}{T+1} \sum_{t=0}^{T} \mathbb{E}\left( \left\| \nabla \hat{L}(\mathbf{A}, \boldsymbol{\beta}_t) \right\|^2 \right) = 0. \quad (23)$$

*Proof:* Please see Appendix B. ∎

*Theorem 3 (Convergence Performance of 1-Bit GC-DL for Non-Convex Loss Functions with Decaying Learning Rates):* Suppose that the gradients are all bounded by (12), 1-bit GC-DL with decaying learning rates

$$\gamma_t = \frac{1 - \sqrt{1 - 4S \dfrac{\gamma_0 - \gamma_0^2 S}{\sqrt{t+1}}}}{2S}, \gamma_0 < \frac{1}{2S}, \quad (24)$$

converges as follows:

$$\min_{0 \leq t \leq T} E\left( \left\| \nabla \hat{L}(\mathbf{A}, \boldsymbol{\beta}_t) \right\|^2 \right)$$

$$\leq \frac{L^0 - L^*}{(\gamma_0 - \gamma_0^2 S)\sqrt{T+1}}$$

$$\quad (25)$$

$$+ \frac{\gamma_0^2 \left[ 2 + \log(T+1) \right]}{(\gamma_0 - \gamma_0^2 S)\sqrt{T+1}} \frac{w - (1-p)}{1-p} \left( \frac{m-1}{n} + \frac{1}{D} \right) CmS,$$

which indicates that

$$\lim_{T \to +\infty} \min_{0 \leq t \leq T} E\left( \left\| \nabla \hat{L}(\mathbf{A}, \boldsymbol{\beta}_t) \right\|^2 \right) = 0. \quad (26)$$

*Proof:* Please see Appendix C. ∎

## V. NUMERICAL RESULTS

### A. Simulations with strongly convex loss functions

Here, the performance of the proposed method is simulated for a linear regression problem whose loss function is strongly convex. We generate a synthetic training dataset $\mathbf{A} \in \mathbb{R}^{1000 \times 101}$ in the following way. Each row of $\mathbf{A}$ represents a training data sample composed of a data point and a label, i.e., $\mathbf{a}_i = [\mathbf{x}_i, y_i]$ , $\mathbf{x}_i \in \mathbb{R}^{1 \times 100}$ , $y \in \mathbb{R}$ . The elements in $\mathbf{x}_i$ are generated independently from the

normal distribution $\mathcal{N}(0,100)$ . The labels in $\mathbf{A}$ are produced as

$$y_i = \langle \mathbf{x}_i, \boldsymbol{\beta}^* \rangle + a_{\text{noise},i}, \quad (27)$$

where the elements in $\boldsymbol{\beta}^* \in \mathbb{R}^{100}$ are generated independently from the standard normal distribution and $a_{\text{noise},i}$ denotes the observation noise drawn from the standard normal distribution. The loss function is defined as

$$\hat{L}(\mathbf{A}, \boldsymbol{\beta}) = \sum_{i=1}^{m} \frac{1}{2} \left( \langle \mathbf{x}_i, \boldsymbol{\beta}^* \rangle - y_i \right)^2. \quad (28)$$

Suppose there are a total number of $n = 100$ workers in the system. In the data distribution stage, each training data sample $\mathbf{a}_i$ is distributed to $d_i$ workers uniformly at random, to approximate the pair-wise balanced scheme, as done in [22]. In the learning stage, in each iteration, unless specified, the probability of each worker to be a straggler is $p = 0.1$ . The parameter vector $\boldsymbol{\beta}_i$ is initialized as a random vector whose elements are randomly and independently drawn from the standard normal distribution. The learning rate is set as $\gamma_t = 0.00001/t$ .

First, we compare the learning performance of different methods, including 1-bit GC-DL, SGC-DL and Ignore-stragglers 1-bit DL. Here Ignore-stragglers 1-bit DL allocates the training data samples to the workers without redundancy, where 1-bit information is transmitted by the workers as done in [33, 34]. This method can be realized by setting $d_i = 1, \forall i$ in the proposed method. For the proposed method and SGC-DL, we fix $d_i = 20$ . In Fig. 2(a), $\sqrt{2\hat{L}(\mathbf{A}, \boldsymbol{\beta}_t)}$ is plotted as a function of the communication overhead[3], where the communication overhead in $t$ iterations is defined as

$$\psi_t \triangleq t\rho, \quad (29)$$

and $\rho$ denotes the number of bits transmitted by each non-straggler worker in each iteration. In Fig. 2(b), error $\left\| \boldsymbol{\beta}_t - \boldsymbol{\beta}^* \right\|$ is plotted as a function of the communication overhead $\psi_t$ . In Fig. 2(c) and (d), $\sqrt{2\hat{L}(\mathbf{A}, \boldsymbol{\beta}_t)}$ and $\left\| \boldsymbol{\beta}_t - \boldsymbol{\beta}^* \right\|$ are plotted as functions of the number of iterations, respectively. From Fig. 2(a) and (b), it is evident that the proposed method attains better learning performance under the same communication overhead compared with the baseline methods. The reason is stated as follows. In each iteration, 1-bit GC-DL requires each non-straggler worker to transmit only 164 bits, whereas SGC-DL requires transmission of 6400 bits per non-straggler worker. Although the proposed method necessitates a higher number of iterations in comparison to SGC-DL, which is implied by Fig. 2(c) and (d), the overall communication overhead is reduced moderately by the proposed method due to significant reduction in communication burden per iteration. In other words, when the communication overhead is fixed, increasing the number of iterations, as done in our method, has a more favorable impact on learning performance compared to increasing the precision of the information transmitted in a single iteration, as done in SGC-DL. It is important to emphasize that even this moderate reduction in communication overhead is highly valuable, particularly in real-world applications where communication resources are often scarce. A moderate decrease in communication overhead can lead to reduced time delays in communication and enhanced training efficiency. Compared with Ignore-stragglers 1-bit DL, the proposed method performs better, albeit at the cost of adding redundancy to the training data distribution. This trade-off is intuitive. In practice, communication resources often pose a greater bottleneck than local computation demands and storage. This suggests that adopting 1-bit GC-DL is both economical and advantageous.

---

[3] Reducing communication overhead throughout the training process is crucial in real-world scenarios, where limitations are often imposed by restricted bandwidth. This is the rationale behind our decision to analyze the performance of the proposed method primarily in terms of communication overhead rather than the number of iterations or energy consumption, as done in much prior work such as [28, 40].

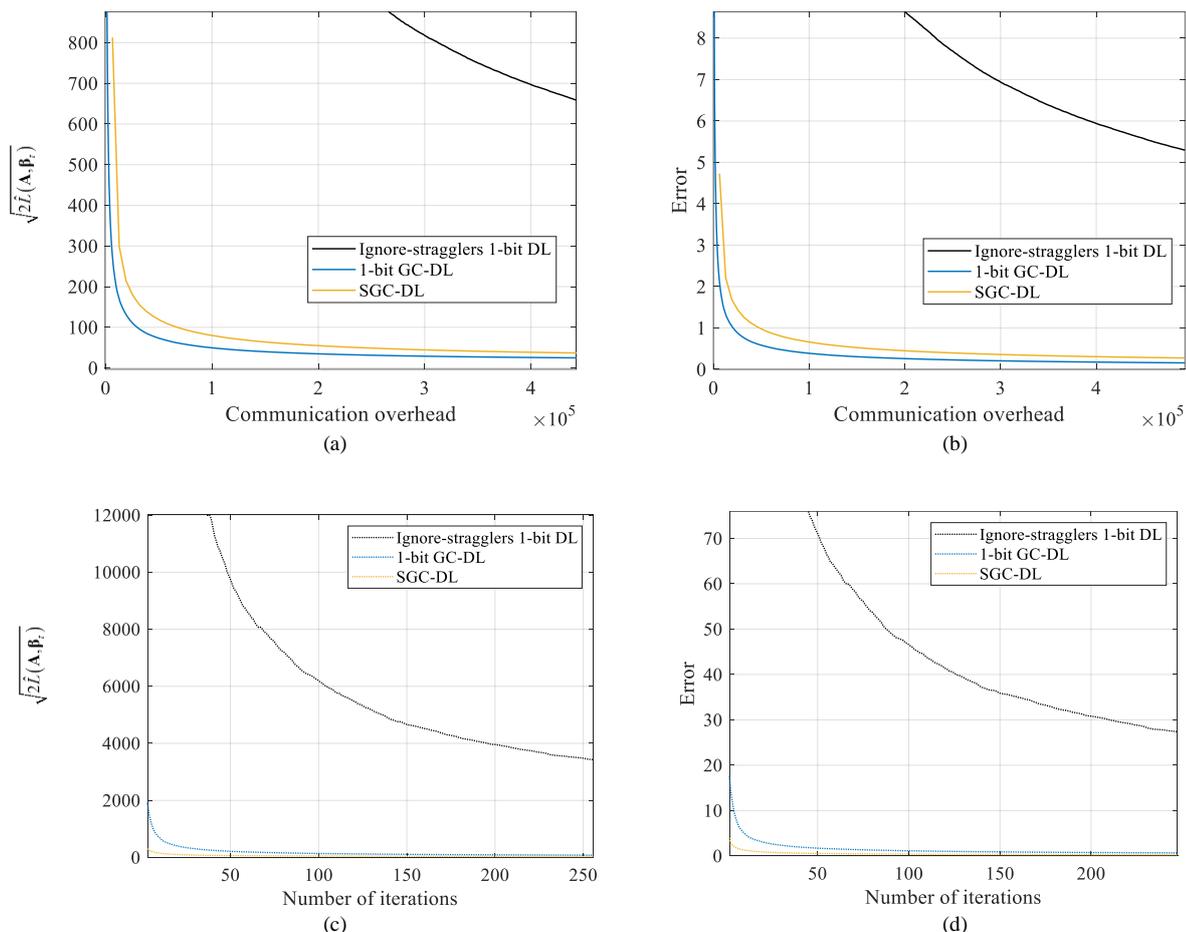

Fig. 2. Learning performance of 1-bit GC-DL, SGC-DL and Ignore-stragglers 1-bit DL, where we set $d_i = 20, \forall i$ in 1-bit GC-DL and SGC-DL. (a) $\sqrt{2\hat{L}(\mathbf{A}, \boldsymbol{\beta}_t)}$ plotted as a function of the communication overhead. (b) $\|\boldsymbol{\beta}_t - \boldsymbol{\beta}^*\|$ plotted as a function of the communication overhead. (c) $\sqrt{2\hat{L}(\mathbf{A}, \boldsymbol{\beta}_t)}$ plotted as a function of the number of iterations. (d) $\|\boldsymbol{\beta}_t - \boldsymbol{\beta}^*\|$ plotted as a function of the number of iterations.

Next, the optimality of the homogeneous data distribution scheme is verified. To this end, in Fig. 3, we compare the learning performance of the proposed method under different values of $\{d_i, \forall i\}$, where the average redundancy of data distribution defined in (16) is fixed at $D = 15$. It can be easily observed from Fig. 3 that the proposed method converges faster in the homogeneous data distribution scheme when $d_i = 15, \forall i$, which is consistent with the theoretical results derived in Section IV.A.

In Fig. 4(a) and (b), to verify that the learning performance of the proposed method improves with increasing average redundancy of data distribution, as indicated in Section IV.A, we plot $\sqrt{2\hat{L}(\mathbf{A}, \boldsymbol{\beta}_t)}$ and $\|\boldsymbol{\beta}_t - \boldsymbol{\beta}^*\|$ as functions of the communication overhead under different values of $D$ in the homogeneous data distribution scheme. It can be seen that the proposed method converges faster with an increasing value of $D$, which is consistent with the theoretical results in Section IV.A. However, faster convergence of the proposed method with increasing average redundancy of data distribution is at the cost of higher computation load. This can be clearly observed from Fig. 4(c) and (d), which show the trade-off between communication overhead and average redundancy of data distribution under the same training loss and the same error, respectively. In practice, a trade-off between computation load induced by redundant distribution of the training data samples and convergence performance should be taken into careful consideration.

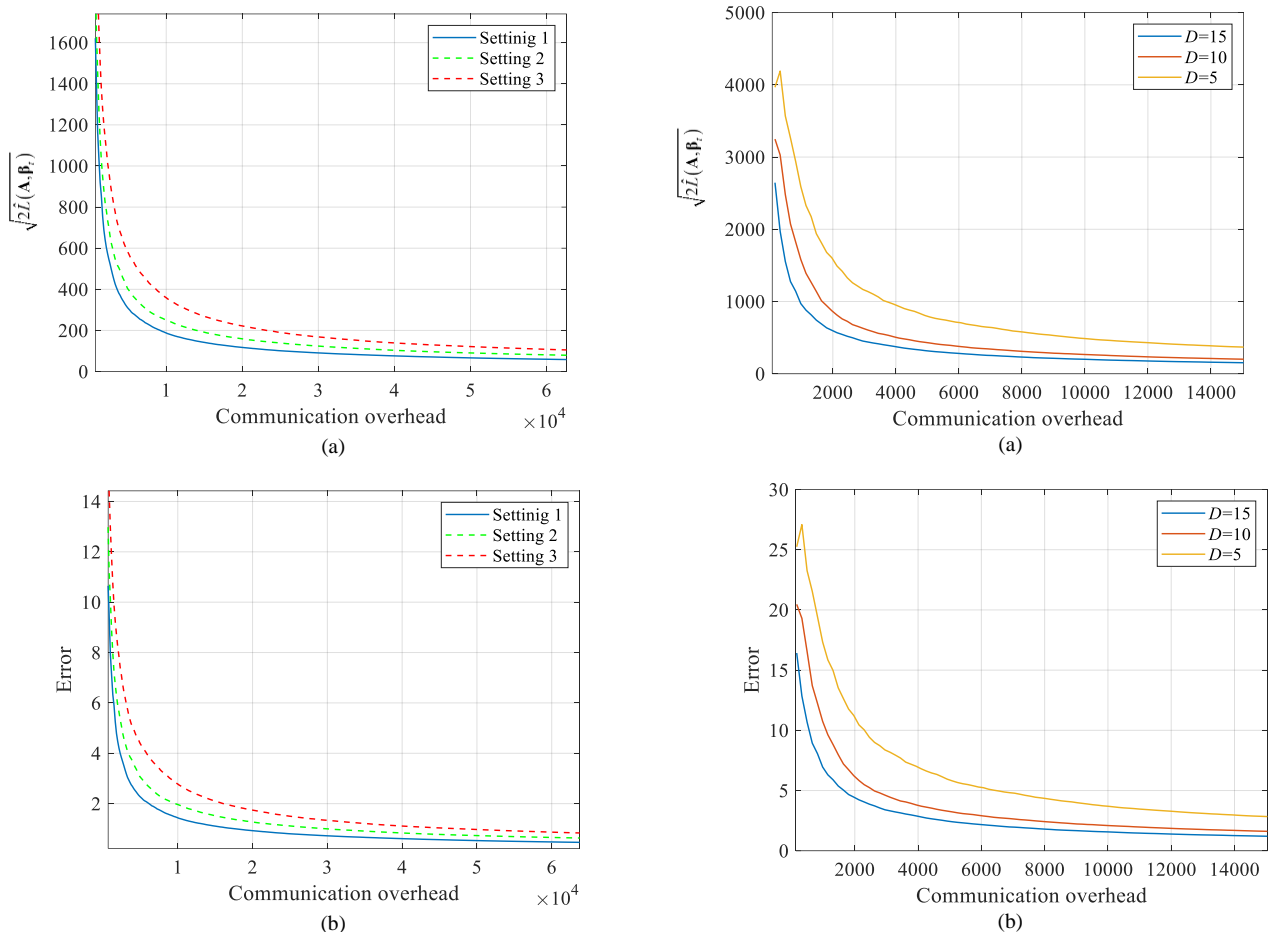

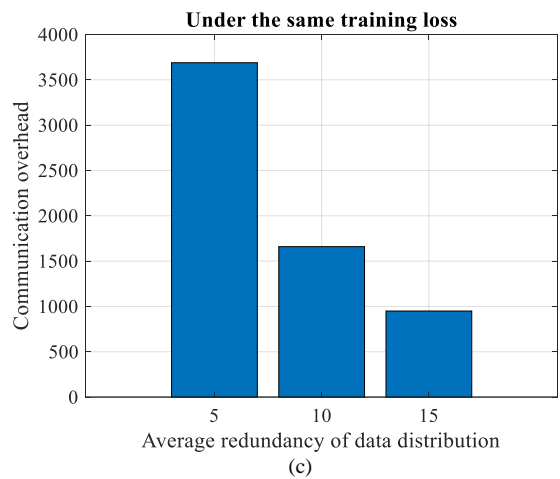

Fig. 3. Learning performance of 1-bit GC-DL under different values of $\{d_i, \forall i\}$, where we set $D = 15$. Under setting 1, we have $d_i = 15, \forall i$. Under setting 2, we have $d_i = 10, i \le m / 2$ and $d_i = 20, i > m / 2$. Under setting 3, we have $d_i = 5, i \le m / 2$ and $d_i = 25, i > m / 2$. (a) $\sqrt{2\hat{L}(\mathbf{A}, \boldsymbol{\beta}_t)}$ plotted as a function of the communication overhead. (b) $\left\| \boldsymbol{\beta}_t - \boldsymbol{\beta}^* \right\|$ plotted as a function of the communication overhead.

From (15), we can observe that a smaller value of $p$ indicates faster convergence of the proposed method. To verify this, in Fig. 5, we plot $\sqrt{2\hat{L}(\mathbf{A}, \boldsymbol{\beta}_t)}$ and $\left\| \boldsymbol{\beta}_t - \boldsymbol{\beta}^* \right\|$ as functions of the communication overhead under different values of $p$ in the homogeneous data distribution scheme, where we fix $D = 15$. It is evident that superior learning performance is achieved when $p$ is reduced. This aligns with intuitive expectations. Even with the redundant distribution of training data across workers, reducing the number of stragglers ensures that more information is provided by the workers in each iteration, leading to a more informed update of the parameter vector and thus, improved learning performance.

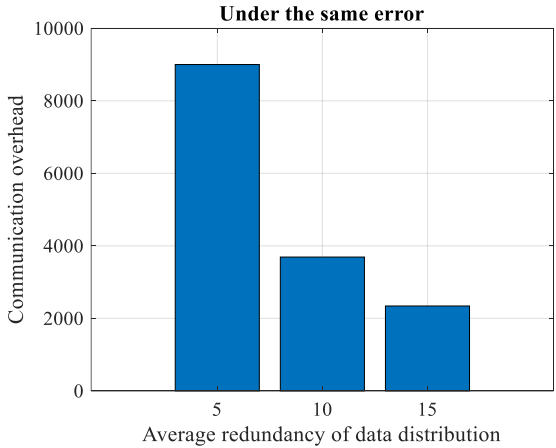

Fig. 4. Learning performance of 1-bit GC-DL under different values of $D$ in the homogeneous data distribution scheme. (a) $\sqrt{2\hat{L}(\mathbf{A},\boldsymbol{\beta}_t)}$ plotted as a function of the communication overhead; (b) $\|\boldsymbol{\beta}_t - \boldsymbol{\beta}^*\|$ plotted as a function of the communication overhead. (c) Trade-off between communication overhead and average redundancy of data distribution under the same training loss, where $\sqrt{2\hat{L}(\mathbf{A},\boldsymbol{\beta}_t)} = 1000$ . (d) Trade-off between communication overhead and average redundancy of data distribution under the same error, where $\|\boldsymbol{\beta}_t - \boldsymbol{\beta}^*\| = 4$ .

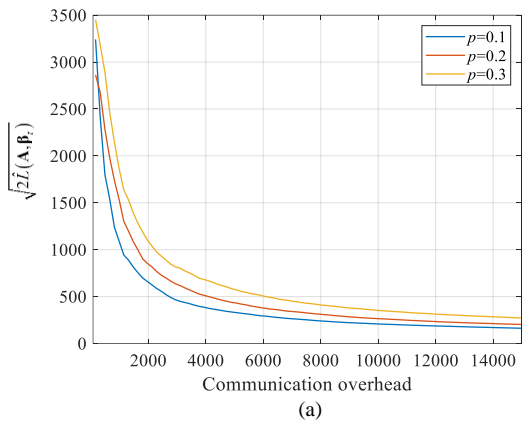

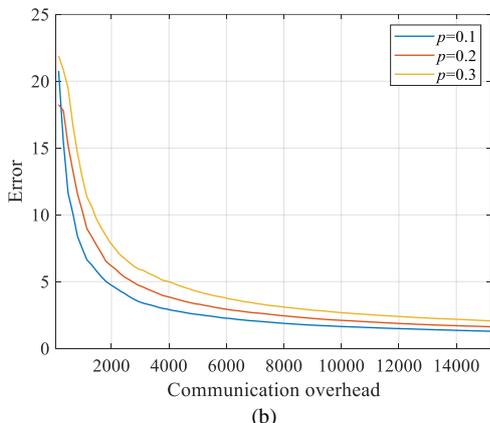

Fig. 5. Learning performance of 1-bit GC-DL under different values of $p$ in the homogeneous data distribution scheme, where $D = 15$ . (a) $\sqrt{2\hat{L}(\mathbf{A},\boldsymbol{\beta}_t)}$ plotted as a function of the communication overhead. (b) $\|\boldsymbol{\beta}_t - \boldsymbol{\beta}^*\|$ plotted as a function of the communication overhead.

### B. Simulations with non-convex loss functions

Next, the performance of the proposed method is simulated when the loss function is non-convex. Here we use the well-known Rosenbrock function [34, 36] as the loss function, which can be expressed as

$$\hat{L}(\mathbf{A},\boldsymbol{\beta}) = \sum_{i=1}^{m} L(\mathbf{a}_i,\boldsymbol{\beta}) = \sum_{i=1}^{m} 100\left(\beta_{i+1} - \beta_i^2\right)^2 + \left(1 - \beta_i\right)^2, \quad (30)$$

where $m = 1000$ and $\boldsymbol{\beta} = \left[\beta_1, ..., \beta_{1001}\right]^T \in \mathbb{R}^{1001}$ . There are a total number of $n = 100$ workers in the system. In the data distribution stage, each training data sample $\mathbf{a}_i$ is distributed to $d_i = 10$ workers, which implies $D = 10$ . The parameter vector $\boldsymbol{\beta}_t$ is initialized as a random vector and its elements are randomly and independently drawn from the standard normal distribution. The learning rate is fixed at $\gamma_t = \gamma = 0.00001$ . In Fig. 6, we plot the training loss as a function of the communication overhead for 1-bit GC-DL and SGC-DL with varying probabilities of each worker to be a straggler. We have excluded the results for Ignore-stragglers 1-bit DL due to its convergence failure. Fig. 6 shows that the proposed method enhances learning performance compared to SGC-DL, achieving faster convergence to the optimal point. Moreover, our method surpasses the performance of SGC-DL even when the probability of each worker to be a straggler is higher for the former than the latter. This underscores the robustness of our approach to the straggler behavior of workers.

To further investigate the sensitivity of the proposed method to the system parameters, in Fig. 7, we plot the training loss as a function of the communication overhead under varying values of learning rate, the number of workers and the average redundancy of data distribution $D$ . It can be seen that the performance of the proposed method remains almost unchanged under slight variations of the system parameters, which demonstrates its robustness.

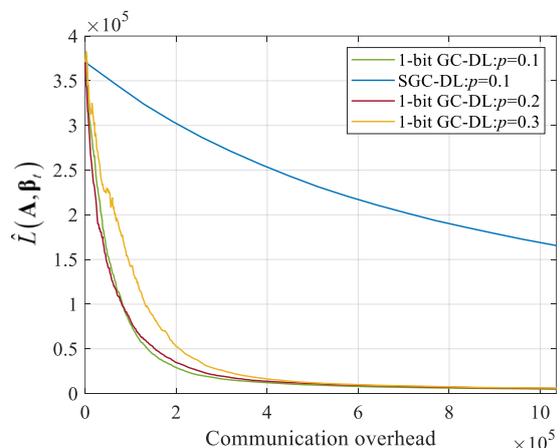

Fig. 6. Training loss of 1-bit GC-DL and SGC-DL as a function of the communication overhead under varying values of $p$ .

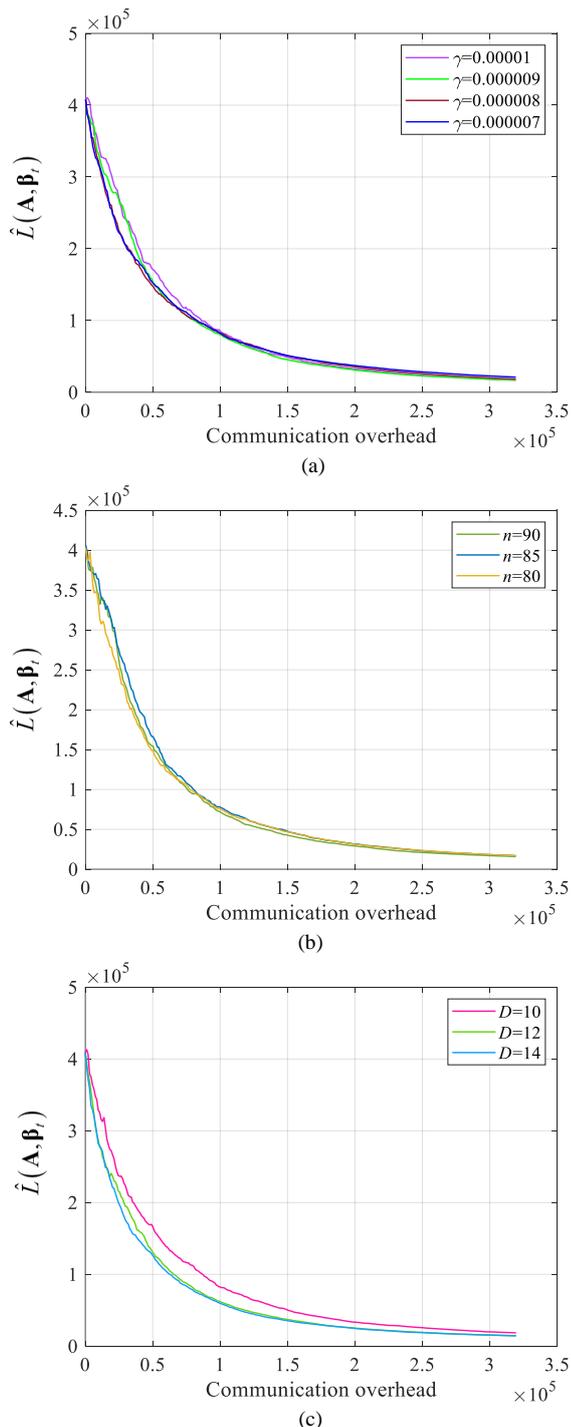

(a)

(b)

(c)

Fig. 7. Training loss of 1-bit GC-DL as a function of the communication overhead under varying system parameters. (a) Under varying values of learning rate. (b) Under varying values of number of workers. (c) Under varying values of average redundancy of data distribution.

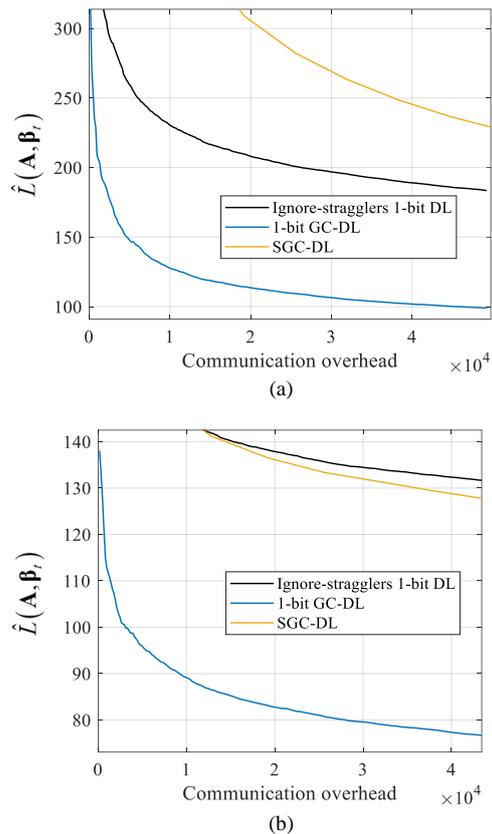

(a)

(b)

Fig. 8. Learning performance of 1-bit GC-DL, SGC-DL and Ignore-stragglers 1-bit DL on real-world datasets. (a) Training loss plotted as a function of the communication overhead on MNIST dataset. (b) Training loss plotted as a function of the communication overhead on Fashion-MNIST dataset.

### C. Experiments on real-world datasets

To further demonstrate the performance of the proposed method, we conduct experiments on two real-world datasets: the MNIST dataset [48] and the Fashion-MNIST dataset [49]. For the MNIST dataset, $m = 100$ training data samples are randomly selected from the training set, representing digits '0' and '2'. These samples are labeled as '-1' and '+1', respectively. Similarly, for the Fashion-MNIST dataset, $m = 100$ training data samples are randomly selected to represent two types of clothing: 'T-shirt/top' and 'Trouser', with labels set to '-1' and '+1', respectively. The reason for using a subset of 100 training data samples from two classes in the entire dataset is to demonstrate the superiority of our proposed method within a limited training time. It is important to note that the performance trends observed with these 100 training data samples are consistent with those seen when using more training data samples from more classes. This consistency implies that our findings in this paper are representative and can be generalized to broader cases. For each dataset, a logistic regression classifier is built to classify these two classes. The loss function is defined as

$$\hat{L}(\mathbf{A}, \boldsymbol{\beta}) = \sum_{i=1}^{m} L(\mathbf{a}_i, \boldsymbol{\beta}) = \sum_{i=1}^{m} \log\left(1 + e^{-y_i \langle \boldsymbol{\beta}, \mathbf{x}_i \rangle}\right), \quad (31)$$

where $\mathbf{a}_i = [\mathbf{x}_i, y_i]$, $\mathbf{x}_i$ denotes the $i$-th training data point that has been vectorized, $y_i$ is the label and $\boldsymbol{\beta}$ is the parameter vector. The system comprises $n = 10$ workers. In the data distribution stage of the proposed method and SGC-DL, each training data point and its corresponding label are distributed at random to $d_i = 2$ workers, approximating the pair-wise balanced scheme described in [22]. During the learning stage, the probability of each worker to be a straggler is fixed at $p = 0.1$. The parameter vector is initialized as a random vector, with elements independently drawn from a standard normal distribution. The learning rate is set to $\gamma_t = 0.001/t$. In Ignore-stragglers 1-bit DL, the training data samples are distributed randomly and uniformly to the workers without redundancy. Fig. 8 compares the learning performance of different methods by plotting the training loss as a function of communication overhead, and it is evident that the proposed method achieves better learning performance compared to baseline methods. This demonstrates that, although the proposed method adopts inaccurate transmission based on 1-bit quantization, the resulting transmission inaccuracies can be offset by increasing the number of iterations. This approach maintains the learning performance while significantly reducing the overall communication overhead.

## VI. Conclusions

In this paper, we considered the problem of DL in the presence of stragglers. Noting that the existing DL methods based on GC require the workers to transmit real-valued vectors during the process of learning and induce very high communication burden, we proposed 1-bit GC-DL, where the communication overhead can be largely reduced by letting the workers to transmit 1-bit data encoded from the locally computed gradients instead of real-valued vectors. The convergence guarantees of the proposed method for both the convex loss functions and non-convex loss functions were provided from a theoretical perspective. It was shown by numerical results that 1-bit GC-DL attains better learning performance under the same communication overhead, compared with the baseline methods. In the future, we plan to extend the proposed method to incorporate other quantization techniques, such as multi-bit quantization techniques and those in [38, 39, 43], to attain a better trade-off between the learning performance and communication overhead. In addition, we will apply the proposed method in different specific applications to achieve an optimal trade-off between communication and computation load by minimizing the overall energy consumption based on the framework provided in this paper.

## Appendix A

Based on (8), $\mathbb{E}\left(\left\|\hat{\mathbf{g}}_r\right\|^2 \middle| \mathcal{F}_{0:t}\right)$ can be expressed as

$$
\begin{aligned}
\mathbb{E}\left(\left\|\hat{\mathbf{g}}_r\right\|^2 \middle| \mathcal{F}_{0:t}\right) &= \mathbb{E}\left(\left\|\sum_{j=1}^{n} I'_j \mathbf{h}'_j \left\|\mathbf{f}'_j\right\|\right\|^2 \middle| \mathcal{F}_{0:t}\right) \\
&= \mathbb{E}\left(\left\langle \sum_{j=1}^{n} I'_j \mathbf{h}'_j \left\|\mathbf{f}'_j\right\|, \sum_{j=1}^{n} I'_j \mathbf{h}'_j \left\|\mathbf{f}'_j\right\|\right\rangle \middle| \mathcal{F}_{0:t}\right) \\
&= \mathbb{E}\left(\sum_{j_1=1}^{n} \sum_{j_2=1}^{n} I'_{j_1} I'_{j_2} \left\langle \mathbf{h}'_{j_1}, \mathbf{h}'_{j_2}\right\rangle \left\|\mathbf{f}'_{j_1}\right\| \left\|\mathbf{f}'_{j_2}\right\| \middle| \mathcal{F}_{0:t}\right) \\
&= \sum_{j_1=1}^{n} \sum_{j_2=1}^{n} \left\|\mathbf{f}'_{j_1}\right\| \left\|\mathbf{f}'_{j_2}\right\| \mathbb{E}\left[I'_{j_1} I'_{j_2} \left\langle \mathbf{h}'_{j_1}, \mathbf{h}'_{j_2}\right\rangle \middle| \mathcal{F}_{0:t}\right].
\end{aligned}
\tag{32}
$$

In (32), if $j_1 = j_2$, based on (6), we have

$$
\begin{aligned}
\mathbb{E}\left[I'_{j_1} I'_{j_1} \left\langle \mathbf{h}'_{j_1}, \mathbf{h}'_{j_1}\right\rangle \middle| \mathcal{F}_{0:t}\right] &= \mathbb{E}\left[I'_{j_1} \left\langle \mathbf{h}'_{j_1}, \mathbf{h}'_{j_1}\right\rangle \middle| \mathcal{F}_{0:t}\right] \\
&= \mathbb{E}\left[I'_{j_1} \middle| \mathcal{F}_{0:t}\right] \mathbb{E}\left[\left\|\mathbf{h}'_{j_1}\right\|^2 \middle| \mathcal{F}_{0:t}\right] = (1-p) w.
\end{aligned}
\tag{33}
$$

Here the second equality is due to the independence between the straggler behavior of the workers and the random 1-bit quantization. If $j_1 \neq j_2$, we have

$$
\begin{aligned}
&\mathbb{E}\left[I'_{j_1} I'_{j_2} \left\langle \mathbf{h}'_{j_1}, \mathbf{h}'_{j_2}\right\rangle \middle| \mathcal{F}_{0:t}\right] \\
&= (1-p)^2 \mathbb{E}\left[\left\langle \mathbf{h}'_{j_1}, \mathbf{h}'_{j_2}\right\rangle \middle| \mathcal{F}_{0:t}\right] \\
&= (1-p)^2 \\
&\times \sum_{k=1}^{w} \left\{\left(\frac{1}{2} + \frac{\left(\mathbf{f}'_{j_1}\right)_k}{2\left\|\mathbf{f}'_{j_1}\right\|}\right)\left(\frac{1}{2} + \frac{\left(\mathbf{f}'_{j_2}\right)_k}{2\left\|\mathbf{f}'_{j_2}\right\|}\right) + \left(\frac{1}{2} - \frac{\left(\mathbf{f}'_{j_1}\right)_k}{2\left\|\mathbf{f}'_{j_1}\right\|}\right)\left(\frac{1}{2} - \frac{\left(\mathbf{f}'_{j_2}\right)_k}{2\left\|\mathbf{f}'_{j_2}\right\|}\right) \right. \\
&\left. - \left(\frac{1}{2} + \frac{\left(\mathbf{f}'_{j_1}\right)_k}{2\left\|\mathbf{f}'_{j_1}\right\|}\right)\left(\frac{1}{2} - \frac{\left(\mathbf{f}'_{j_2}\right)_k}{2\left\|\mathbf{f}'_{j_2}\right\|}\right) - \left(\frac{1}{2} - \frac{\left(\mathbf{f}'_{j_1}\right)_k}{2\left\|\mathbf{f}'_{j_1}\right\|}\right)\left(\frac{1}{2} + \frac{\left(\mathbf{f}'_{j_2}\right)_k}{2\left\|\mathbf{f}'_{j_2}\right\|}\right) \right\} \\
&= (1-p)^2 \frac{\left\langle \mathbf{f}'_{j_1}, \mathbf{f}'_{j_2}\right\rangle}{\left\|\mathbf{f}'_{j_1}\right\| \left\|\mathbf{f}'_{j_2}\right\|},
\end{aligned}
\tag{34}
$$

where the first equality is due to the independence between the straggler behavior of the workers and the random 1-bit quantization and the independence among the straggler behavior of different workers, the second equality is derived by substituting (6) into (34), and $\left(\mathbf{f}'_j\right)_k$ is the $k$-th element of $\mathbf{f}'_j$.

Next, substituting (33) and (34) into (32), we can derive

$$\mathbb{E}\left(\left\|\hat{\mathbf{g}}_t\right\|^2 \middle| \mathcal{F}_{0:t}\right)$$

$$= \sum_{j_1=1}^{n}\sum_{j_2=1}^{n}\left(1-p\right)^2 \left\langle \mathbf{f}_{j_1}^t, \mathbf{f}_{j_2}^t\right\rangle + \left[\left(1-p\right)w - \left(1-p\right)^2\right]\sum_{j=1}^{n}\left\langle \mathbf{f}_j^t, \mathbf{f}_j^t\right\rangle$$

$$= \sum_{j_1=1}^{n}\sum_{j_2=1}^{n}\left\langle \sum_{i_1\in S_{j_1}}\frac{\nabla L\left(\mathbf{a}_{i_1},\boldsymbol{\beta}_t\right)}{d_{i_1}}, \sum_{i_2\in S_{j_2}}\frac{\nabla L\left(\mathbf{a}_{i_2},\boldsymbol{\beta}_t\right)}{d_{i_2}}\right\rangle$$

$$+ \frac{w-\left(1-p\right)}{1-p}\sum_{j=1}^{n}\left\langle \sum_{i_1\in S_j}\frac{\nabla L\left(\mathbf{a}_{i_1},\boldsymbol{\beta}_t\right)}{d_{i_1}}, \sum_{i_2\in S_j}\frac{\nabla L\left(\mathbf{a}_{i_2},\boldsymbol{\beta}_t\right)}{d_{i_2}}\right\rangle$$

$$= \sum_{i_1=1}^{m}\sum_{i_2=1}^{m}\sum_{j_1=1}^{n}\sum_{j_2=1}^{n} I\left(i_1\in S_{j_1}\right)I\left(i_2\in S_{j_2}\right)$$

$$\times\left\langle \frac{\nabla L\left(\mathbf{a}_{i_1},\boldsymbol{\beta}_t\right)}{d_{i_1}}, \frac{\nabla L\left(\mathbf{a}_{i_2},\boldsymbol{\beta}_t\right)}{d_{i_2}}\right\rangle$$

$$+ \frac{w-\left(1-p\right)}{1-p}\sum_{i_1=1}^{m}\sum_{i_2=1}^{m}\sum_{j=1}^{n} I\left(i_1\in S_j\right)I\left(i_2\in S_j\right)$$

$$\times\left\langle \frac{\nabla L\left(\mathbf{a}_{i_1},\boldsymbol{\beta}_t\right)}{d_{i_1}\left(1-p\right)}, \frac{\nabla L\left(\mathbf{a}_{i_2},\boldsymbol{\beta}_t\right)}{d_{i_2}\left(1-p\right)}\right\rangle. \tag{35}$$

Here the second equality is derived by substituting (5) into (35). In (35), we have

$$\sum_{i_1=1}^{m}\sum_{i_2=1}^{m}\sum_{j_1=1}^{n}\sum_{j_2=1}^{n} I\left(i_1\in S_{j_1}\right)I\left(i_2\in S_{j_2}\right)\left\langle \frac{\nabla L\left(\mathbf{a}_{i_1},\boldsymbol{\beta}_t\right)}{d_{i_1}}, \frac{\nabla L\left(\mathbf{a}_{i_2},\boldsymbol{\beta}_t\right)}{d_{i_2}}\right\rangle$$

$$= \sum_{i_1=1}^{m}\sum_{i_2=1}^{m} d_{i_2}d_{i_1}\left\langle \frac{\nabla L\left(\mathbf{a}_{i_1},\boldsymbol{\beta}_t\right)}{d_{i_1}}, \frac{\nabla L\left(\mathbf{a}_{i_2},\boldsymbol{\beta}_t\right)}{d_{i_2}}\right\rangle$$

$$= \sum_{i_1=1}^{m}\sum_{i_2=1}^{m}\left\langle \nabla L\left(\mathbf{a}_{i_1},\boldsymbol{\beta}_t\right), \nabla L\left(\mathbf{a}_{i_2},\boldsymbol{\beta}_t\right)\right\rangle, \tag{36}$$

where the first equality is derived by noting that

$$\sum_{j_2=1}^{n} I\left(i_2\in S_{j_2}\right) = d_{i_2}, \sum_{j_1=1}^{n} I\left(i_1\in S_{j_1}\right) = d_{i_1}. \tag{37}$$

In (35), we have

$$\sum_{i_1=1}^{m}\sum_{i_2=1}^{m}\sum_{j=1}^{n} I\left(i_1\in S_j\right)I\left(i_2\in S_j\right)\left\langle \frac{\nabla L\left(\mathbf{a}_{i_1},\boldsymbol{\beta}_t\right)}{d_{i_1}\left(1-p\right)}, \frac{\nabla L\left(\mathbf{a}_{i_2},\boldsymbol{\beta}_t\right)}{d_{i_2}\left(1-p\right)}\right\rangle$$

$$= \sum_{i_1=1}^{m}\sum_{i_2=1,i_1\neq i_2}^{m} \frac{d_{i_1}d_{i_2}}{n}\left\langle \frac{\nabla L\left(\mathbf{a}_{i_1},\boldsymbol{\beta}_t\right)}{d_{i_1}}, \frac{\nabla L\left(\mathbf{a}_{i_2},\boldsymbol{\beta}_t\right)}{d_{i_2}}\right\rangle$$

$$+ \sum_{i=1}^{m} d_i\left\langle \frac{\nabla L\left(\mathbf{a}_i,\boldsymbol{\beta}_t\right)}{d_i}, \frac{\nabla L\left(\mathbf{a}_i,\boldsymbol{\beta}_t\right)}{d_i}\right\rangle, \tag{38}$$

where the first equality is derived from

$$\sum_{j=1}^{n} I\left(i_1\in S_j\right)I\left(i_2\in S_j\right) = \frac{d_{i_1}d_{i_2}}{n}, i_1\neq i_2, \tag{39}$$

in the pair-wise balanced manner of the data distribution. Substituting (36) and (38) into (35), we can obtain

$$\mathbb{E}\left(\left\|\hat{\mathbf{g}}_t\right\|^2 \middle| \mathcal{F}_{0:t}\right)$$

$$= \sum_{i_1=1}^{m}\sum_{i_2=1}^{m}\left\langle \nabla L\left(\mathbf{a}_{i_1},\boldsymbol{\beta}_t\right), \nabla L\left(\mathbf{a}_{i_2},\boldsymbol{\beta}_t\right)\right\rangle$$

$$+ \frac{w-\left(1-p\right)}{1-p}\sum_{i_1=1}^{m}\sum_{i_2=1,i_1\neq i_2}^{m} \frac{1}{n}\left\langle \nabla L\left(\mathbf{a}_{i_1},\boldsymbol{\beta}_t\right), \nabla L\left(\mathbf{a}_{i_2},\boldsymbol{\beta}_t\right)\right\rangle \tag{40}$$

$$+ \frac{w-\left(1-p\right)}{1-p}\sum_{i=1}^{m} \frac{1}{d_i}\left\langle \nabla L\left(\mathbf{a}_i,\boldsymbol{\beta}_t\right), \nabla L\left(\mathbf{a}_i,\boldsymbol{\beta}_t\right)\right\rangle$$

$$\leq Cm^2 + \frac{w-\left(1-p\right)}{1-p}\frac{m^2-m}{n}C + \frac{w-\left(1-p\right)}{1-p}C\sum_{i=1}^{m}\frac{1}{d_i},$$

where the last inequality is derived from

$$\left\langle \nabla L\left(\mathbf{a}_{i_1},\boldsymbol{\beta}_t\right), \nabla L\left(\mathbf{a}_{i_2},\boldsymbol{\beta}_t\right)\right\rangle \leq \left\|\nabla L\left(\mathbf{a}_{i_1},\boldsymbol{\beta}_t\right)\right\|\left\|\nabla L\left(\mathbf{a}_{i_2},\boldsymbol{\beta}_t\right)\right\| \leq C, \tag{41}$$

based on the assumption in (12).

Taking full expectations on both sides of (40) yields (13).

## Appendix B

From (20), we have

$$\hat{L}\left(\mathbf{A},\boldsymbol{\beta}_{t+1}\right) = \hat{L}\left(\mathbf{A},\boldsymbol{\beta}_t - \gamma_t\hat{\mathbf{g}}_t\right)$$

$$\leq \hat{L}\left(\mathbf{A},\boldsymbol{\beta}_t\right) - \left\langle \nabla\hat{L}\left(\mathbf{A},\boldsymbol{\beta}_t\right), \gamma_t\hat{\mathbf{g}}_t\right\rangle + S\left\|\gamma_t\hat{\mathbf{g}}_t\right\|^2. \tag{42}$$

Taking the expectation over the random behavior of the stragglers in the $t-$th iteration conditioned on the previous iterations for (42) yields

$$\mathbb{E}\left[\hat{L}\left(\mathbf{A},\boldsymbol{\beta}_{t+1}\right)\middle|\mathcal{F}_{0:t}\right]$$

$$\leq \hat{L}\left(\mathbf{A},\boldsymbol{\beta}_t\right) - \left\langle \nabla\hat{L}\left(\mathbf{A},\boldsymbol{\beta}_t\right), \gamma_t\mathbb{E}\left(\hat{\mathbf{g}}_t\middle|\mathcal{F}_{0:t}\right)\right\rangle + S\mathbb{E}\left(\left\|\gamma_t\hat{\mathbf{g}}_t\right\|^2\middle|\mathcal{F}_{0:t}\right)$$

$$= \hat{L}\left(\mathbf{A},\boldsymbol{\beta}_t\right) - \gamma_t\left\|\nabla\hat{L}\left(\mathbf{A},\boldsymbol{\beta}_t\right)\right\|^2 + \gamma_t^2 S\mathbb{E}\left(\left\|\hat{\mathbf{g}}_t\right\|^2\middle|\mathcal{F}_{0:t}\right), \tag{43}$$

where the equality is derived from Lemma 1. In (43), we have

$$\mathbb{E}\left(\left\|\hat{\mathbf{g}}_t\right\|^2 \middle| \mathcal{F}_{0:t}\right)$$

$$= \left\|\nabla\hat{L}\left(\mathbf{A},\boldsymbol{\beta}_t\right)\right\|^2$$

$$+ \frac{w-\left(1-p\right)}{1-p}\sum_{i_1=1}^{m}\sum_{i_2=1,i_1\neq i_2}^{m} \frac{1}{n}\left\langle \nabla L\left(\mathbf{a}_{i_1},\boldsymbol{\beta}_t\right), \nabla L\left(\mathbf{a}_{i_2},\boldsymbol{\beta}_t\right)\right\rangle$$

$$+ \frac{w-\left(1-p\right)}{1-p}\sum_{i=1}^{m} \frac{1}{D}\left\|\nabla L\left(\mathbf{a}_i,\boldsymbol{\beta}_t\right)\right\|^2$$

$$\leq \left\|\nabla\hat{L}\left(\mathbf{A},\boldsymbol{\beta}_t\right)\right\|^2$$

$$+ \frac{w-\left(1-p\right)}{1-p}\sum_{i_1=1}^{m}\sum_{i_2=1,i_1\neq i_2}^{m} \frac{1}{n}\frac{\left\|\nabla L\left(\mathbf{a}_{i_1},\boldsymbol{\beta}_t\right)\right\|^2 + \left\|\nabla L\left(\mathbf{a}_{i_2},\boldsymbol{\beta}_t\right)\right\|^2}{2}$$

$$+ \frac{w-\left(1-p\right)}{1-p}\sum_{i=1}^{m} \frac{1}{D}\left\|\nabla L\left(\mathbf{a}_i,\boldsymbol{\beta}_t\right)\right\|^2$$

$$= \left\|\nabla\hat{L}\left(\mathbf{A},\boldsymbol{\beta}_t\right)\right\|^2$$

$$+ \frac{w-\left(1-p\right)}{1-p}\left(\frac{m-1}{n} + \frac{1}{D}\right)\sum_{i=1}^{m}\left\|\nabla L\left(\mathbf{a}_i,\boldsymbol{\beta}_t\right)\right\|^2, \tag{44}$$

where the first equality is obtained from (40) and the inequality is derived by noting $\langle \mathbf{x}, \mathbf{y} \rangle \leq \left( \|\mathbf{x}\|^2 + \|\mathbf{y}\|^2 \right)/2$.

Then, substituting (44) into (43) yields

$$\mathbb{E}\left[ \hat{L}(\mathbf{A}, \boldsymbol{\beta}_{t+1}) \Big| \mathcal{F}_{0:t} \right]$$
$$\leq \hat{L}(\mathbf{A}, \boldsymbol{\beta}_t) + \left( \gamma_t^2 S - \gamma_t \right) \left\| \nabla \hat{L}(\mathbf{A}, \boldsymbol{\beta}_t) \right\|^2 \qquad (45)$$
$$+ \gamma_t^2 S \frac{w - (1-p)}{1-p} \left( \frac{m-1}{n} + \frac{1}{D} \right) \sum_{i=1}^m \left\| \nabla L(\mathbf{a}_i, \boldsymbol{\beta}_t) \right\|^2,$$

which is equivalent to

$$\left( \gamma_t - \gamma_t^2 S \right) \left\| \nabla \hat{L}(\mathbf{A}, \boldsymbol{\beta}_t) \right\|^2$$
$$\leq \hat{L}(\mathbf{A}, \boldsymbol{\beta}_t) - \mathbb{E}\left[ \hat{L}(\mathbf{A}, \boldsymbol{\beta}_{t+1}) \Big| \mathcal{F}_{0:t} \right]$$
$$+ \gamma_t^2 S \frac{w - (1-p)}{1-p} \left( \frac{m-1}{n} + \frac{1}{D} \right) \sum_{i=1}^m \left\| \nabla L(\mathbf{a}_i, \boldsymbol{\beta}_t) \right\|^2.$$
$$(46)$$

Taking full expectations on both sides of (46), we can obtain

$$\left( \gamma_t - \gamma_t^2 S \right) \mathbb{E}\left( \left\| \nabla \hat{L}(\mathbf{A}, \boldsymbol{\beta}_t) \right\|^2 \right)$$
$$\leq \mathbb{E}\left[ \hat{L}(\mathbf{A}, \boldsymbol{\beta}_t) \right] - \mathbb{E}\left[ \hat{L}(\mathbf{A}, \boldsymbol{\beta}_{t+1}) \right]$$
$$+ \gamma_t^2 S \frac{w - (1-p)}{1-p} \left( \frac{m-1}{n} + \frac{1}{D} \right) \sum_{i=1}^m \mathbb{E}\left( \left\| \nabla L(\mathbf{a}_i, \boldsymbol{\beta}_t) \right\|^2 \right) \qquad (47)$$
$$\leq \mathbb{E}\left[ \hat{L}(\mathbf{A}, \boldsymbol{\beta}_t) \right] - \mathbb{E}\left[ \hat{L}(\mathbf{A}, \boldsymbol{\beta}_{t+1}) \right]$$
$$+ \gamma_t^2 S \frac{w - (1-p)}{1-p} \left( \frac{m-1}{n} + \frac{1}{D} \right) Cm,$$

where the second inequality is derived from (12). Based on (47), we have

$$\sum_{t=0}^T \left( \gamma_t - \gamma_t^2 S \right) \mathbb{E}\left( \left\| \nabla \hat{L}(\mathbf{A}, \boldsymbol{\beta}_t) \right\|^2 \right)$$
$$\leq \sum_{t=0}^T \left\{ \mathbb{E}\left[ \hat{L}(\mathbf{A}, \boldsymbol{\beta}_t) \right] - \mathbb{E}\left[ \hat{L}(\mathbf{A}, \boldsymbol{\beta}_{t+1}) \right] \right\}$$
$$+ S \frac{w - (1-p)}{1-p} \left( \frac{m-1}{n} + \frac{1}{D} \right) Cm \sum_{t=0}^T \gamma_t^2$$
$$= \hat{L}(\mathbf{A}, \boldsymbol{\beta}_0) - \mathbb{E}\left[ \hat{L}(\mathbf{A}, \boldsymbol{\beta}_{T+1}) \right] \qquad (48)$$
$$+ S \frac{w - (1-p)}{1-p} \left( \frac{m-1}{n} + \frac{1}{D} \right) Cm \sum_{t=0}^T \gamma_t^2$$
$$\leq L^0 - L^* + S \frac{w - (1-p)}{1-p} \left( \frac{m-1}{n} + \frac{1}{D} \right) Cm \sum_{t=0}^T \gamma_t^2,$$

where the last inequality is derived based on the fact that $\mathbb{E}\left[ \hat{L}(\mathbf{A}, \boldsymbol{\beta}_{T+1}) \right] \geq L^*$.

If the learning rate is fixed as (21) with $T > \left( 4S \right)^{4/3} - 1$, we have

$$\gamma - \gamma^2 S = \frac{1}{\left( T+1 \right)^{3/4}} > 0. \qquad (49)$$

Combining (48) and (49), we have

$$\frac{1}{T+1} \sum_{t=0}^T \mathbb{E}\left( \left\| \nabla \hat{L}(\mathbf{A}, \boldsymbol{\beta}_t) \right\|^2 \right)$$
$$\leq \frac{L^0 - L^*}{\left( \gamma - \gamma^2 S \right)(T+1)} + S \frac{w - (1-p)}{1-p} \left( \frac{m-1}{n} + \frac{1}{D} \right) Cm \frac{\gamma^2}{\gamma - \gamma^2 S}$$
$$= \frac{L^0 - L^*}{\left( T+1 \right)^{1/4}}$$
$$+ S \frac{w - (1-p)}{1-p} \left( \frac{m-1}{n} + \frac{1}{D} \right) Cm \left( \frac{1 - \sqrt{1 - \dfrac{4S}{\left( T+1 \right)^{3/4}}}}{2S} \right)^2 \left( T+1 \right)^{3/4}.$$
$$(50)$$

Note that, the following limit holds:

$$\lim_{T \to \infty} \left( T+1 \right)^{3/4} \left( \frac{1 - \sqrt{1 - 4 \dfrac{S}{\left( T+1 \right)^{3/4}}}}{2S} \right)^2$$
$$(51)$$
$$= \lim_{\gamma \to 0^+} \frac{\gamma^2}{\gamma - \gamma^2 S} = \lim_{\gamma \to 0^+} \frac{2\gamma}{1 - 2\gamma S} = 0,$$

which completes the proof.

## APPENDIX C

With the learning rates given in (24), we have

$$\gamma_t - \gamma_t^2 S = \frac{\gamma_0 - \gamma_0^2 S}{\sqrt{t+1}} > 0. \qquad (52)$$

Based on the derivations (42)-(48) in Appendix B and (52), the following holds for the considered case:

$$\sum_{t=0}^T \frac{\gamma_0 - \gamma_0^2 S}{\sqrt{t+1}} \mathbb{E}\left( \left\| \nabla \hat{L}(\mathbf{A}, \boldsymbol{\beta}_t) \right\|^2 \right)$$
$$\leq L^0 - L^* + S \frac{w - (1-p)}{1-p} \left( \frac{m-1}{n} + \frac{1}{D} \right) Cm \sum_{t=0}^T \gamma_t^2. \qquad (53)$$

Based on (53), we have

$$\min_{0 \leq t \leq T} \mathbb{E}\left( \left\| \nabla \hat{L}(\mathbf{A}, \boldsymbol{\beta}_t) \right\|^2 \right) \leq \frac{\sum_{t=0}^T \dfrac{\gamma_0 - \gamma_0^2 S}{\sqrt{t+1}} \mathbb{E}\left( \left\| \nabla \hat{L}(\mathbf{A}, \boldsymbol{\beta}_t) \right\|^2 \right)}{\sum_{t=0}^T \dfrac{\gamma_0 - \gamma_0^2 S}{\sqrt{t+1}}}$$
$$\leq \frac{L^0 - L^* + S \dfrac{w - (1-p)}{1-p} \left( \dfrac{m-1}{n} + \dfrac{1}{D} \right) Cm \sum_{t=0}^T \gamma_t^2}{\sum_{t=0}^T \dfrac{\gamma_0 - \gamma_0^2 S}{\sqrt{t+1}}}$$
$$= \frac{L^0 - L^*}{\sum_{t=0}^T \dfrac{\gamma_0 - \gamma_0^2 S}{\sqrt{t+1}}}$$
$$+ \frac{\sum_{t=0}^T \gamma_t^2}{\left( \gamma_0 - \gamma_0^2 S \right) \sum_{t=0}^T \dfrac{1}{\sqrt{t+1}}} \frac{w - (1-p)}{1-p} \left( \frac{m-1}{n} + \frac{1}{D} \right) Cm S$$

$$\leq \frac{L^0 - L^*}{(\gamma_0 - \gamma_0^2 S)\sqrt{T+1}}$$

$$+ \frac{\sum_{t=0}^{T} \frac{\gamma_0^2}{t+1}}{(\gamma_0 - \gamma_0^2 S)\sqrt{T+1}} \frac{w-(1-p)}{1-p}\left(\frac{m-1}{n}+\frac{1}{D}\right)CmS$$

$$\leq \frac{L^0 - L^*}{(\gamma_0 - \gamma_0^2 S)\sqrt{T+1}}$$

$$+ \frac{\gamma_0^2\left[2+\log(T+1)\right]}{(\gamma_0 - \gamma_0^2 S)\sqrt{T+1}} \frac{w-(1-p)}{1-p}\left(\frac{m-1}{n}+\frac{1}{D}\right)CmS, \quad (54)$$

where the last two inequalities are derived from the standard inequalities

$$\sum_{t=0}^{T}\frac{1}{\sqrt{t+1}} \geq \sqrt{T+1}, \sum_{t=0}^{T}\frac{1}{t+1} \leq 2+\log(T+1), \quad (55)$$

and by noting from (24) that

$$(1-\gamma_0 S)\gamma_t \leq \gamma_t - \gamma_t^2 S = \frac{\gamma_0 - \gamma_0^2 S}{\sqrt{t+1}}$$

$$\Rightarrow \gamma_t \leq \frac{\gamma_0}{\sqrt{t+1}} \Rightarrow \sum_{t=0}^{T}\gamma_t^2 \leq \sum_{t=0}^{T}\frac{\gamma_0^2}{t+1}. \quad (56)$$

It can be easily shown that

$$\lim_{T\to\infty}\frac{L^0 - L^*}{(\gamma_0 - \gamma_0^2 S)\sqrt{T+1}} = 0,$$

$$\lim_{T\to\infty}\frac{\gamma_0^2\left[2+\log(T+1)\right]}{(\gamma_0 - \gamma_0^2 S)\sqrt{T+1}} \frac{w-(1-p)}{1-p}\left(\frac{m-1}{n}+\frac{1}{D}\right)CmS = 0,$$

$$(57)$$

which completes the proof.